\documentclass[a4paper]{cas-dc}

\usepackage[numbers]{natbib}

\usepackage{adjustbox}
\usepackage{float}
\usepackage[ruled]{algorithm2e}
\usepackage{algorithmic}
\def\tsc#1{\csdef{#1}{\textsc{\lowercase{#1}}\xspace}}
\tsc{WGM}
\tsc{QE}
\tsc{EP}
\tsc{PMS}
\tsc{BEC}
\tsc{DE}

\begin{document}
\let\WriteBookmarks\relax
\def\floatpagepagefraction{1}
\def\textpagefraction{.001}

\shorttitle{IoT-Driven 3D Pose Optimization}

\shortauthors{LuPing Dai}

\title [mode = title]{IoT-Based 3D Pose Estimation and Motion Optimization for Athletes: Application of C3D and OpenPose}                      








\author[1]{Fei Ren }
\ead{renfei24321@outlook.com}
\cormark[1]

\author[2]{Chao Ren}
\ead{l15716341220@163.com}

\author[3]{Tianyi Lyu}
\ead{lyu.t.iot@gmail.com}

\address[1]{Yong In University,Yong In,17902,Korea}
\address[2]{Jiao Zuo University, Jiao Zuo,Henan,454150,China}

\address[3]{Granite Telecommunications LLC. 100 Newport Avenue Extension, Quincy, MA,02171.USA}





\begin{abstract}
This study proposes the IoT-Enhanced Pose Optimization Network (IE-PONet) for high-precision 3D pose estimation and motion optimization of track and field athletes. IE-PONet integrates C3D for spatiotemporal feature extraction, OpenPose for real-time keypoint detection, and Bayesian optimization for hyperparameter tuning. Experimental results on NTURGB+D and FineGYM datasets demonstrate superior performance, with AP\(^p50\) scores of 90.5 and 91.0, and mAP scores of 74.3 and 74.0, respectively. Ablation studies confirm the essential roles of each module in enhancing model accuracy. IE-PONet provides a robust tool for athletic performance analysis and optimization, offering precise technical insights for training and injury prevention. Future work will focus on further model optimization, multimodal data integration, and developing real-time feedback mechanisms to enhance practical applications.

\end{abstract}

\begin{keywords}
Human Pose Estimation \sep IoT Sensors \sep Motion Capture \sep Multi-View Pose Estimation \sep Deep Learning \sep Athlete Action
\end{keywords}
\maketitle

\section{Introduction}

The development of IoT technology offers new avenues for real-time data collection and analysis. IoT sensors can collect various types of data from athletes' training and competitions in real-time, including video, acceleration, angular velocity, and more. This data can be transmitted wirelessly to a central processing system for analysis, enabling immediate feedback and optimization of athletes' movements. Recently, increasing research has explored the integration of IoT and sports science, particularly in real-time monitoring and feedback. Lu et al. applied IoT technology in track and field, achieving real-time pose monitoring and motion analysis, significantly improving training efficiency and effectiveness. IoT technology is also widely used in sports training, such as real-time monitoring of athletes' physiological indicators through wearable devices, further optimizing training plans and strategies.

The application of three-dimensional human pose estimation and motion optimization in track and field is receiving more and more attention, which can not only significantly improve the technical movement and sports performance of athletes, but also effectively reduce the risk of sports injury. However, with the development of modern sports science, the demand for real-time and accurate posture analysis and feedback is increasing among athletes and coaches. However, the existing methods and technologies still face many challenges and limitations in their practical application.

Although deep learning and computer vision technology in recent years has made significant progress in 3 d human pose estimation, such as Wang et al. proposed 3 d human pose estimation method based on CNN on multiple benchmark datasets showed high robustness and accuracy~\cite{wang2021deep,de2023performance,wang2024using,wang2023computational,liu2025eitnet,weng2024leveraging}, Badiola et al. fusion images 3D pose estimation system improves the accuracy and reliability of pose estimation~\cite{badiola2021systematic}, but most of these methods rely on large-scale offline data processing, difficult to meet the requirements of real-time monitoring and feedback in actual scenarios such as track and field. In addition, the traditional attitude estimation method usually requires a large amount of computing resources and a long processing time, which is more inadequate in the highly dynamic and complex motion environment. This computational delay prevents athletes and coaches from obtaining immediate movement analysis results during training, and thus from making real-time movement adjustment and improvement. In addition, most existing methods still have low computational efficiency when dealing with complex motion scenarios and high-frequency data, which limits their application in actual motion scenarios~\cite{wang2022smart}.

The development of IoT technology has provided new opportunities for real-time motion data acquisition and analysis. With IoT sensors deployed on training sites and athletes, various types of data, such as video, acceleration and angular speed, can be collected in real time, and transmitted to a central processing system in real time over a wireless network for analysis. Although existing studies, such as Lu et al., applied the Internet of Things technology in track and field to realize real-time posture monitoring and movement analysis of athletes, and greatly improved the training efficiency and effect~\cite{lu2021iot,gong2024graphicalstructurallearningrsfmri,Wang2024Theoretical,liu2025real,wang2022classification,peng2024maxk}, these systems still have limitations in dealing with large-scale data transmission, complex environmental adaptability and real-time.

In view of the above challenges, this study proposes the IE-PONet (IoT-Enhanced Pose Optimization Network) model, designed to address these current technical and application challenges. By integrating the advantages of C3D and OpenPose, the IE-PONet model realizes the accurate real-time analysis and optimization of athletes' movements; and dynamically adjusts the hyperparameters of the model through Bayesian optimization to further improve the adaptability and efficiency of the model. Most importantly, IE-PONet realizes the real-time acquisition and transmission of data through the Internet of Things sensor, which not only greatly improves the timeliness of data processing, but also effectively solves the shortcomings of the existing methods in real-time feedback and complex scenarios.

The main contributions of this paper are as follows:
\begin{itemize}
  \item [$\bullet$]
Proposing the IE-PONet model, designed specifically for motion analysis and optimization of track and field athletes by combining C3D, OpenPose, and Bayesian optimization.
  \item [$\bullet$]
Achieving effective integration of IoT and deep learning. Real-time data collection and transmission through IoT sensors are combined with deep learning models for real-time analysis and feedback.
  \item [$\bullet$]
Validation on FineGYM and NTURGB+D datasets. Extensive experiments validate the superior performance and robustness of the IE-PONet model on different datasets.
\end{itemize}

The structure of this paper is as follows: Section 2 reviews related research work, including 3D human pose estimation, the application of IoT in sports analysis, and optimization techniques in deep learning. Section 3 introduces the overall architecture and various modules of the IE-PONet model in detail. Section 4 describes the experimental setup, datasets, experimental environment, and performance metrics, followed by a detailed analysis of the experimental results. Section 5 summarizes the research work, discussing its limitations and future research directions.

\section{RELATED WORK}

In recent years, significant progress has been made in 3D human pose estimation, the application of IoT in sports analysis, and optimization techniques in deep learning, driving the development of athlete motion analysis and optimization. The following sections provide a detailed review of these related works.

\subsection{3D Human Pose Estimation}

In the field of three-dimensional human pose estimation, although significant progress has been made in recent years, especially in the application of convolutional neural network (CNN) and long and short-term memory network (LSTM), these methods still have some shortcomings and challenges. Although the method based on multi-scale convolutional network proposed by xu et al. achieves high precision attitude estimation by integrating multi-level feature information, its method relies on complex network structure and has high computational cost, making it difficult to maintain high efficiency in real-time applications~\cite{xu2021graph, lee2024traffic,peng2024automatic,zhuang2020music,Shen2024Harnessing,yuan2025gta,weng2024big,zhang2024deep,zhang2024cu,shi2019multi,chen2024enhancing}. Similarly, the pose estimation model that integrates spatial and temporal features developed by Zhang et al. appears inadequate in response to high-frequency movements, despite its superior performance on multiple benchmark datasets, because the method may experience lag or estimation bias in dealing with fast movement changes~\cite{zhang2023hierarchical}.

Zou et al.'s graph convolution network (GCN) method performs well in handling the key points of the human body, and can better capture the structural information of the human body, but its adaptability to multiple perspectives is limited, especially vulnerable to occlusion, light changes and other factors in complex scenes~\cite{zou2021modulated, luo2023fleet,an2023runtime,wang2024deep,liu2024design,qiao2024av4ev,wang2024intelligent,liu2024dsem,chen2024few,wang2018performance,huang2024risk}. And multiple perspective method while effectively improve the accuracy of pose estimation, such as sun et als~\cite{sun2020multi}. Using three-dimensional pose reconstruction, and Remelli et al proposed multiple perspective deep learning model, enhance the robustness of the model, but these methods often need multiple perspectives of data input, which in practical application by equipment and environmental conditions, it is difficult to widely promote~\cite{remelli2020lightweight}. Although the self-supervised learning method proposed by Srivastav et al. reduces the dependence on labeled data, self-supervised learning may perform less than those with labeled data when handling complex scenarios or high-speed movements~\cite{srivastav2024selfpose3d, sui2024application,peng2023autorep,li2024deep,li2024optimizing,richardson2024reinforcement,zhou2024optimization,wan2024image,yang2019regional}.

Although existing methods have made remarkable progress in three-dimensional human pose estimation, there face challenges in handling real-time, high-frequency motion, as well as complex scenarios. The IE-PONet model proposed in this paper is improved for these problems. By combining the advantages of C3D and OpenPose, and adopting Bayesian optimization to improve the accuracy and efficiency of the model, so as to better meet the challenges in these practical applications. We believe that more robust and more efficient solutions can be achieved in the field of pose estimation by further optimizing the model structure and methods.




\subsection{Applications of IoT in Sports Analysis}

The application of IoT technology in sports analysis has been increasing in recent years, improving the timeliness and accuracy of sports monitoring and analysis through real-time data collection and transmission. Li et al. demonstrated the application of IoT sensors in real-time motion capture, proposing an IoT-based sports monitoring system that enables real-time data collection and analysis of athletes' training~\cite{li2021real}. Meanwhile, Kofahi et al. developed a real-time motion analysis platform using IoT technology, providing instant feedback during training and significantly enhancing training effectiveness~\cite{a2021smart}. The combination of IoT and deep learning offers new solutions for sports analysis. Luo et al.~\cite{luo2020full} proposed a real-time pose estimation method that combines IoT and deep learning, achieving efficient data processing and analysis through edge computing~\cite{wang2024recording,wang2024cross,zhu2024complex,cao2018expected,xi2024enhancing,dong2024design,caoapplication,li2024ltpnet,wang2021machine,xu2022dpmpc,weng2024fortifying}. Rajavel et al. further optimized this method by introducing a distributed computing architecture, enhancing system scalability and processing capability~\cite{rajavel2022iot}. Additionally, Deepa et al. demonstrated the application of IoT in sports injury prevention, providing personalized prevention advice through real-time monitoring and data analysis~\cite{deepa2023iot}. However, existing research mainly focuses on offline data processing, failing to meet the needs for real-time monitoring and feedback. The IE-PONet model proposed in this paper addresses this gap by achieving real-time data collection and transmission through IoT sensors, combined with deep learning models for real-time analysis and feedback.

Despite the promising application prospects of IoT technology in motion analysis, the existing methods still face many challenges in real-time performance, data fusion, processing speed, and feedback accuracy~\cite{ning1, gong2020research,de2022modeling,chen2024mix,jin2023visual,qiao2024robust,jiang2020dualvd,zhou2024adapi,zheng2024identification,tang2024real,yan2024application}. The IE-PONet model proposed in this paper can provide more efficient and accurate real-time monitoring and feedback in complex motion scenarios by integrating the real-time data acquisition and transmission capabilities of the Internet of Things sensors, as well as the real-time analysis and optimization capabilities of the deep learning model. This innovation not only compensates for the shortcomings of existing research, but also provides new perspectives and approaches for future potential in broader motion analysis and optimization applications.


\subsection{Optimization Techniques in Deep Learning}

Optimization techniques play a crucial role in the training of deep learning models. In recent years, Bayesian optimization has gained widespread application due to its superior performance in hyperparameter tuning. Cho et al. achieved automated tuning of deep learning models through Bayesian optimization, significantly improving model performance~\cite{cho2020basic}. Additionally, Zhang et al. proposed a hybrid optimization method that combines Bayesian optimization and evolutionary algorithms, demonstrating outstanding performance in multiple tasks~\cite{zhang2021convolutional}. Besides Bayesian optimization, other optimization techniques are also widely used in deep learning. Luo et al. studied the application of the Stochastic Gradient Descent (SGD) algorithm on large-scale datasets, proposing a new momentum adjustment strategy that improved model convergence speed and stability~\cite{luo2020efficient}. Meanwhile, Tang et al. further optimized the training process of deep learning models by introducing an adaptive learning rate optimization algorithm~\cite{tang2021improved}. In recent years, hybrid methods combining multiple optimization techniques have shown remarkable performance in enhancing deep learning models. Mahendran et al. proposed a hybrid method combining genetic algorithms and Bayesian optimization, validating its superior performance across different tasks through experiments~\cite{mahendran2021improving}. Additionally, Lingam et al. demonstrated a method combining particle swarm optimization and deep reinforcement learning, successfully applied to solving complex tasks~\cite{lingam2020particle}. Despite significant achievements in enhancing the performance of deep learning models, optimization and tuning of models in practical applications still pose challenges. The IE-PONet model proposed in this paper enhances the accuracy and efficiency of pose estimation and motion analysis through Bayesian optimization for hyperparameter tuning, addressing some of the issues present in existing methods~\cite{ning2}.



\section{METHODOLOGY}

\subsection{Overview of Our Network}

\begin{figure*}[ht]
    \centering
    \includegraphics[width=0.9\textwidth]{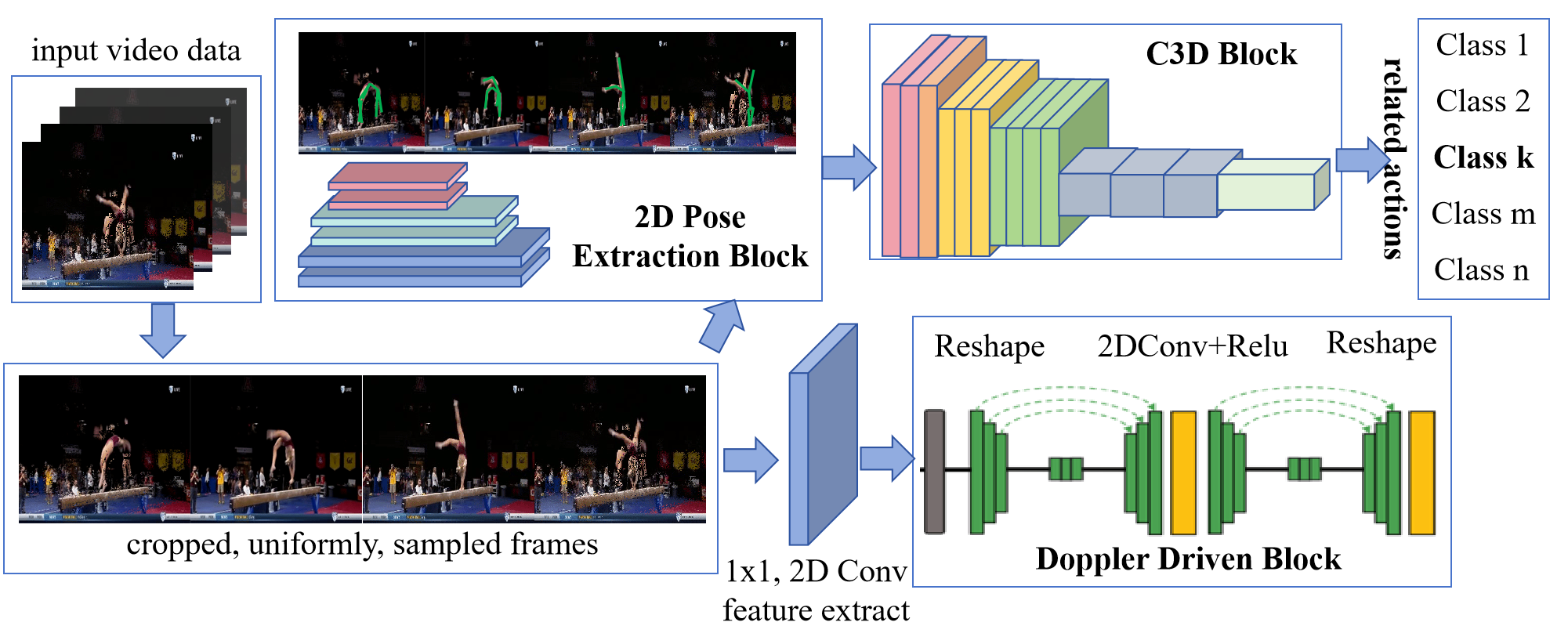}  
    \caption{Overall flow chart of IE-PONet Model Structure.} 
    \label{fig-overview}
\end{figure*}

The structure of the IE-PONet (IoT-Enhanced Pose Optimization Network) model proposed in this paper is shown in Figure~\ref{fig-overview}. This model achieves real-time analysis and optimization of track and field athletes' movements by combining C3D, OpenPose, and Bayesian optimization.

Video data of the athletes is collected in real-time by IoT sensors and transmitted to the system for processing. After preprocessing, the data is input into the C3D (Convolutional 3D Network) module. The C3D module is responsible for capturing the video data features of the athletes' movements, extracting spatiotemporal features from the video through three-dimensional convolution operations. This helps retain the dynamic information of the athletes' movements, improving the model's sensitivity to changes in motion.

The extracted spatiotemporal features are passed to the OpenPose module for pose estimation. The OpenPose module estimates poses by detecting key points of the human body in real-time. This module processes each frame of the video using convolutional neural networks to extract the 2D coordinates of human key points and generates 3D pose information of the athletes through feature fusion. This stage provides high-precision pose data for subsequent motion analysis and optimization.

To further improve the model's performance, the Bayesian optimization module is introduced for hyperparameter optimization. Bayesian optimization explores and adjusts the hyperparameter space to find the best parameter combination for optimal model performance, thus enhancing the accuracy and efficiency of pose estimation and motion analysis. In this paper, Bayesian optimization is used not only to tune the hyperparameters of C3D and OpenPose but also for the overall optimization of the entire IE-PONet model.

After obtaining the 3D pose information of the athletes, the system analyzes and classifies the data. By combining multi-layer convolutional neural networks (3D CNN), the model classifies and identifies the athletes' movements. The 3D CNN module utilizes the previously extracted spatiotemporal features and pose information to accurately classify the athletes' movements, identify different action categories, and provide optimization suggestions.

The IE-PONet model achieves real-time analysis and optimization of athletes' movements by organically integrating C3D, OpenPose, and Bayesian optimization, utilizing IoT technology to significantly enhance training effectiveness and athletic performance. Figure 11 illustrates the overall structure of the IE-PONet model and the relationships between its modules.








\subsection{C3D Module}

\begin{figure}[ht]
    \centering
	\includegraphics[width=0.45\textwidth]{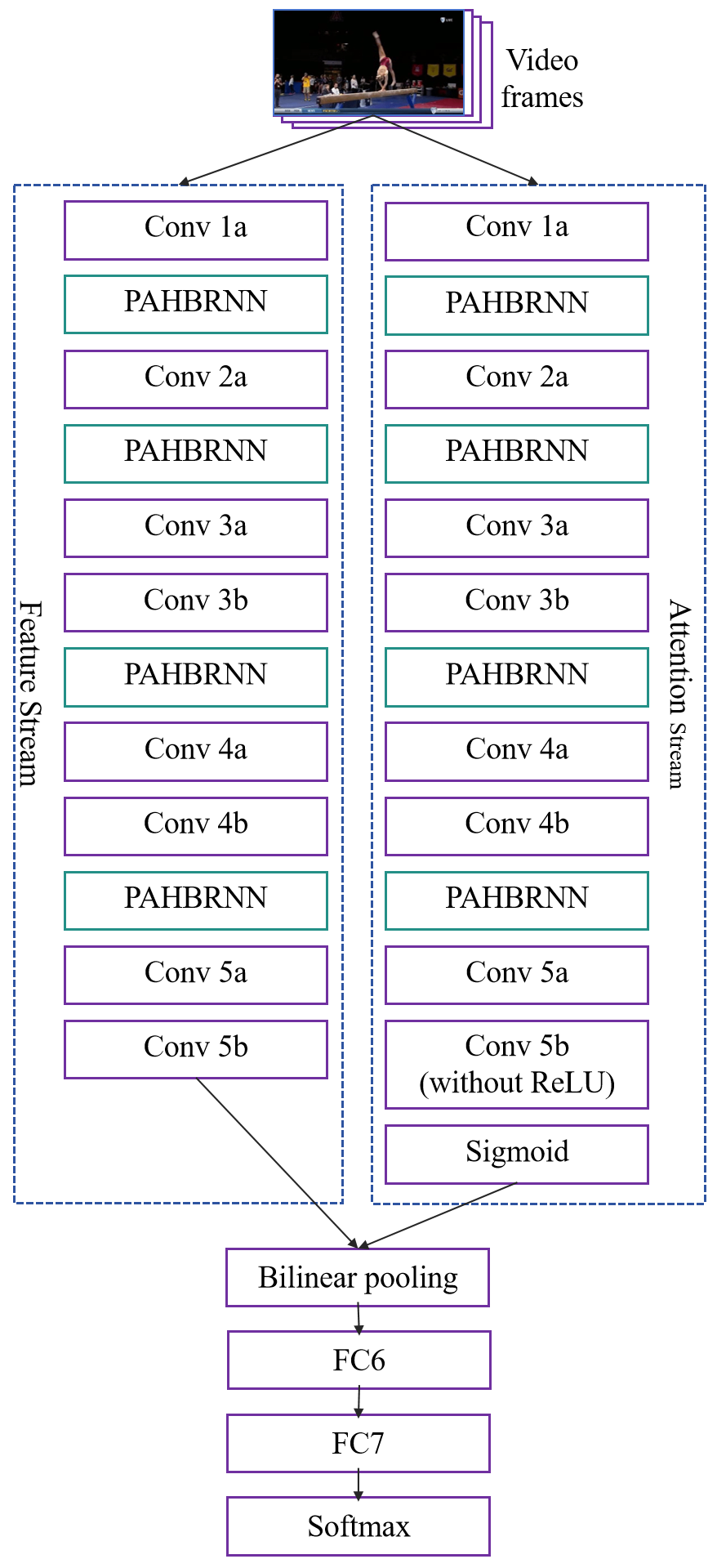}  
	\caption{Flow chart of the C3D Module Structure.} 
	\label{fig-2}
\end{figure}

In the IE-PONet model, the C3D (Convolutional 3D Network) module is responsible for capturing the video data features of athletes' movements. Compared to traditional 2D convolutional neural networks (2D CNNs), C3D has a stronger capability to extract spatiotemporal features, as it can capture both spatial and temporal information from the video~\cite{joo2021exemplar, gao2021dynamic}. This makes C3D perform better in handling continuous movements and dynamic changes. While 2D CNNs can only process single frames and cannot capture the continuity of movements, C3D, through three-dimensional convolution operations, processes each frame in the video sequence and integrates information from adjacent frames, thereby providing a more complete and accurate representation of the movement features. Figure~\ref{fig-2} shows the structural flowchart of the C3D module.

The C3D module first preprocesses the input video frames, converting them into a format suitable for convolution operations. Video data is typically input as a series of frames, each with a fixed time interval. The preprocessing stage includes normalization and resizing of video frames to ensure consistency and efficiency in subsequent processing. The normalization process involves scaling the pixel values of video frames to a fixed range, which improves the stability and speed of model training. Resizing adjusts the video frames to a fixed size to meet the input requirements of the convolutional network. In Figure~\ref{fig-2}, the preprocessed video frames enter the first 3D convolutional layer. By sliding the 3D convolutional kernel across spatial and temporal dimensions, the C3D module can extract spatiotemporal features from the video. These features capture the continuous changes and dynamic information of the athletes' movements, providing richer feature representation than 2D convolutional networks. The key equations in the C3D module are as follows:

Input Preprocessing:

\begin{equation}
X_{pre} = \text{Preprocess}(X)
\end{equation}
where \( X \) represents the raw video data and \( X_{pre} \) represents the preprocessed video data.

3D Convolution Operation:
\begin{equation}
X_{conv1} = \sigma(W_{conv1} * X_{pre} + b_{conv1})
\end{equation}
where \( * \) denotes the convolution operation, \( W_{conv1} \) and \( b_{conv1} \) are the first layer convolution kernel and bias, \( \sigma \) denotes the activation function (e.g., ReLU), and \( X_{conv1} \) is the feature map after convolution.

Pooling Operation:
\begin{equation}
X_{pool1} = \text{Pool}(X_{conv1})
\end{equation}
where \(\text{Pool}\) denotes the pooling operation (e.g., max pooling or average pooling), and \( X_{pool1} \) is the feature map after pooling.

Multi-layer Convolution and Pooling:
\begin{equation}
X_{convN} = \sigma(W_{convN} * X_{pool(N-1)} + b_{convN})
\end{equation}

\begin{equation}
X_{poolN} = \text{Pool}(X_{convN})
\end{equation}
where \( N \) denotes the N-th convolutional and pooling layers, \( W_{convN} \) and \( b_{convN} \) are the convolution kernel and bias for the N-th layer, and \( X_{convN} \) and \( X_{poolN} \) are the feature maps after N-th convolution and pooling layers.

Feature Fusion:
\begin{equation}
X_{fused} = \text{Fuse}(X_{pool1}, X_{pool2}, \ldots, X_{poolN})
\end{equation}
where \(\text{Fuse}\) denotes the feature fusion operation, which integrates feature maps from multiple pooling layers into a comprehensive feature map, \( X_{fused} \).

Spatiotemporal Feature Extraction:
\begin{equation}
X_{spatial\_temporal} = \sigma(W_{st} * X_{fused} + b_{st})
\end{equation}
where \( W_{st} \) and \( b_{st} \) are the convolution kernel and bias for the spatiotemporal feature extraction layer, and \( X_{spatial\_temporal} \) is the extracted spatiotemporal feature map.

Feature Output:
\begin{equation}
X_{out} = \text{Output}(X_{spatial\_temporal})
\end{equation}
where \(\text{Output}\) denotes the output layer operation, which converts the spatiotemporal feature map into the final feature output, \( X_{out} \).

In the structure of the C3D module, in addition to the basic convolutional and pooling layers, bilinear pooling is employed for feature fusion between the feature system and the attention system. Bilinear pooling, as an efficient feature fusion technique, captures high-order interactions between different feature maps, thereby enhancing the richness and discriminative power of feature representations. Bilinear pooling also improves the robustness of the model. In complex movement scenarios, different motion details may share similar low-order features, but their high-order features often exhibit significant differences. By utilizing bilinear pooling, the C3D module can more effectively distinguish these high-order features, thus improving the ability to differentiate between various motion patterns. In this module, bilinear pooling is used to integrate the output features from both the feature system and the attention system, enabling the model to consider both global information and local details of the action features, enhancing its adaptability and stability in complex movement scenarios.

As shown in Figure~\ref{fig-2}, the bilinear pooling operation is placed between the outputs of the feature system and the attention system. By applying bilinear pooling to the output feature maps of these two systems, a comprehensive feature representation is generated, which is then fed into fully connected layers (FC6 and FC7) for further feature processing and classification.

\begin{equation}
X_{bilinear} = (X_{feature} \otimes X_{attention})
\end{equation}
where \( X_{feature} \) and \( X_{attention} \) represent the output feature maps from the feature system and the attention system, respectively, and \( \otimes \) denotes the outer product operation. In this way, the model can fully utilize the diverse information from both feature sources, providing a more accurate foundation for subsequent classification and pose estimation.

From the preprocessing of input video frames, through the 3D convolution and pooling operations, to the multi-layer convolution and pooling feature extraction, and finally through feature fusion and spatiotemporal feature extraction, the C3D module efficiently captures motion features in the video, providing rich information for subsequent pose estimation and motion analysis.





\subsection{OpenPose Module}

\begin{figure}[ht]
    \centering
    \includegraphics[width=0.5\textwidth]{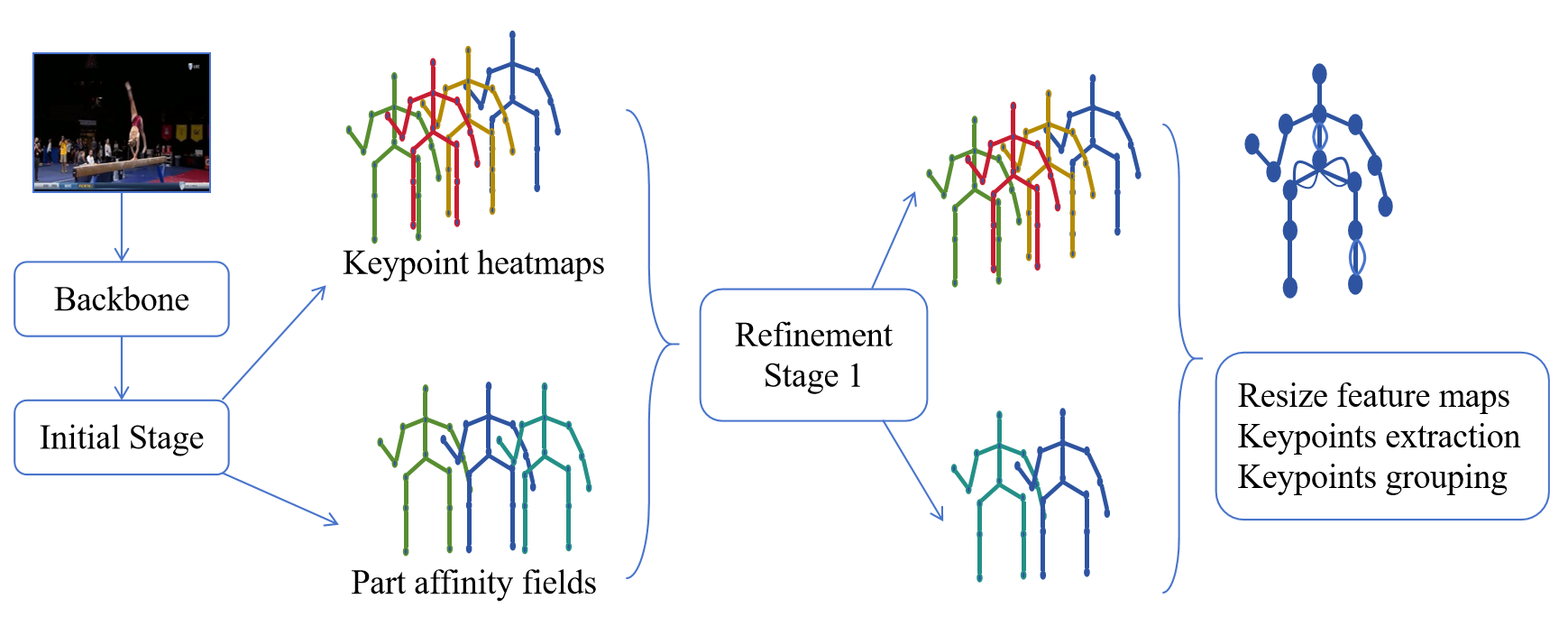}  
    \caption{Structural Flow chart of the OpenPose Module.} 
    \label{fig-3}
\end{figure}

In the IE-PONet model, the OpenPose module is responsible for real-time detection of human keypoints for pose estimation. This module processes each frame of the image using convolutional neural networks to extract the 2D coordinates of various keypoints of the human body and generates 3D pose information of the athlete through feature fusion~\cite{kim2021ergonomic, zheng2024triz}. The OpenPose module excels in handling complex movements and pose variations, providing high-precision keypoint detection results~\cite{mroz2021comparing}. Compared with traditional methods, OpenPose can not only detect more keypoints but also maintain high accuracy and robustness in more complex backgrounds. This makes it widely used in the field of motion analysis and pose estimation. Figure~\ref{fig-3} shows the structural flowchart of the OpenPose module.

The OpenPose module first preprocesses the input image data, including image normalization and resizing. The normalization process adjusts the pixel values of the image to a fixed range (usually 0 to 1) to ensure that different image data are processed on the same scale, improving the stability and speed of model training. Resizing scales the input image to a fixed size required by the model, facilitating subsequent convolution operations. This preprocessing step not only eliminates differences in image size but also improves the efficiency of convolution operations. The preprocessed image data is input into the backbone network. The backbone network is the core part of the OpenPose module and is responsible for extracting high-level features from the image. This network typically consists of multiple convolutional and pooling layers, which extract and compress feature maps layer by layer to achieve multi-scale feature extraction from the image. In the first convolution operation, the 3D convolution kernel slides across spatial and temporal dimensions to extract the initial features of the image. Each convolutional layer is accompanied by a nonlinear activation function (e.g., ReLU) to introduce nonlinearity and enhance the expressive power of the model. The key equations in the OpenPose module are as follows:

Input Preprocessing:

\begin{equation}
I_{pre} = \text{Preprocess}(I)
\end{equation}
where \( I \) represents the raw image data, and \( I_{pre} \) represents the preprocessed image data.

First Convolution Operation:
\begin{equation}
F_{conv1} = \sigma(W_{conv1} * I_{pre} + b_{conv1})
\end{equation}
where \( * \) denotes the convolution operation, \( W_{conv1} \) and \( b_{conv1} \) are the first layer convolution kernel and bias, \( \sigma \) denotes the activation function (e.g., ReLU), and \( F_{conv1} \) is the feature map after convolution.

The feature map after the first convolution operation undergoes a pooling operation (e.g., max pooling or average pooling) to reduce the spatial dimensions of the data, thereby lowering computational complexity while retaining important features.

Pooling Operation:
\begin{equation}
F_{pool1} = \text{Pool}(F_{conv1})
\end{equation}
where \(\text{Pool}\) denotes the pooling operation, and \( F_{pool1} \) is the feature map after pooling.

Through multiple layers of convolution and pooling operations, the OpenPose module extracts higher-level feature maps. Each combination of convolution and pooling operations further extracts important features from the image and reduces the size of the feature map. These feature maps contain rich spatial information and partial temporal information, effectively representing the local features and overall structure of the human pose.

Multi-layer Convolution and Pooling:
\begin{equation}
F_{convN} = \sigma(W_{convN} * F_{pool(N-1)} + b_{convN})
\end{equation}

\begin{equation}
F_{poolN} = \text{Pool}(F_{convN})
\end{equation}
where \( N \) denotes the N-th convolutional and pooling layers, \( W_{convN} \) and \( b_{convN} \) are the convolution kernel and bias for the N-th layer, and \( F_{convN} \) and \( F_{poolN} \) are the feature maps after the N-th convolution and pooling layers.

Next, the OpenPose module uses feature fusion technology to merge feature maps from different layers into a comprehensive feature map, providing rich information for keypoint detection. The feature fusion step combines feature maps from multiple layers and integrates them through convolution operations so that the final feature map contains multi-scale information, which can more accurately represent the human pose.

Feature Fusion:
\begin{equation}
F_{fused} = \text{Fuse}(F_{pool1}, F_{pool2}, \ldots, F_{poolN})
\end{equation}
where \(\text{Fuse}\) denotes the feature fusion operation, which merges the feature maps from multiple pooling layers into a comprehensive feature map, \( F_{fused} \).

Based on the data after feature fusion, the OpenPose module performs keypoint detection. Using convolutional layers, it outputs heatmaps of keypoints and Part Affinity Fields (PAFs), which can represent the positions of each keypoint and the connections between adjacent parts.

Keypoint Heatmap:
\begin{equation}
H_{k} = \sigma(W_{heatmap} * F_{fused} + b_{heatmap})
\end{equation}
where \( W_{heatmap} \) and \( b_{heatmap} \) are the convolution kernel and bias for the heatmap convolutional layer, and \( H_{k} \) is the heatmap for the k-th keypoint.

In the initial stage, the initial heatmaps and PAFs of the keypoints are generated. Then, through multiple refinement stages, the positions of the keypoints are continuously optimized, generating more precise heatmaps and PAFs.

Through these operations, the OpenPose module generates heatmaps and PAFs of the keypoints, readjusts the feature maps, and extracts the keypoints, finally grouping the keypoints to generate the 3D pose information of the athlete.

Keypoint Grouping:
\begin{equation}
G_{k} = \text{Group}(H_{k}, P_{aff})
\end{equation}
where \(\text{Group}\) denotes the keypoint grouping operation, and \( G_{k} \) is the grouping result of the k-th keypoint.

The output of the OpenPose module is integrated with the feature output of the C3D module. Through feature fusion technology, it combines spatiotemporal features and keypoint information, providing more comprehensive motion analysis and pose estimation. Through this series of operations, the OpenPose module efficiently detects human keypoints, providing accurate keypoint data for subsequent 3D pose estimation.




\subsection{Bayesian Optimization Module}

\begin{figure}[ht]
    \centering
    \includegraphics[width=0.5\textwidth]{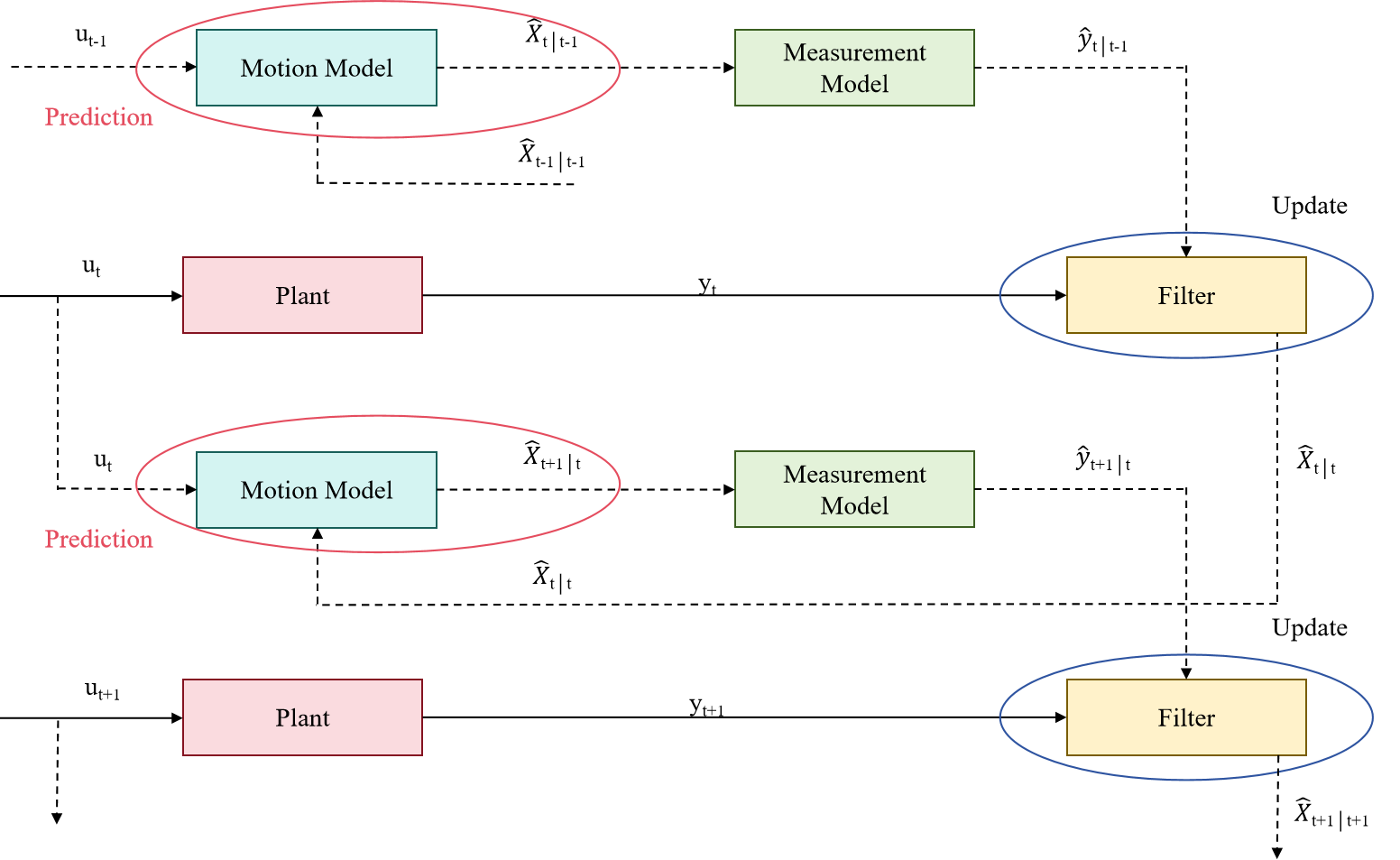}  
    \caption{Structural Flowchart of Bayesian Optimization.} 
    \label{fig-4}
\end{figure}

In the IE-PONet model, the Bayesian optimization module is used to fine-tune the hyperparameters of the model to improve the accuracy and efficiency of pose estimation and motion analysis~\cite{biderman2024lightning,steyvers2022bayesian}. Bayesian optimization is a global optimization method particularly suitable for optimizing black-box functions. Its core principle is to approximate the objective function by constructing a surrogate model and use this surrogate model to guide the selection of hyperparameters. Figure~\ref{fig-4} shows the structural flowchart of Bayesian optimization.

The core idea of Bayesian optimization is to view the hyperparameter optimization problem as a sequential decision-making process. At each step, Bayesian optimization first uses a probabilistic model (usually a Gaussian process) to approximate the objective function. The Gaussian process provides a probabilistic model of the objective function by performing Bayesian inference, which can predict the distribution of the objective function at unobserved points and quantify the uncertainty in the predictions. Based on this, the optimal hyperparameter combination is selected for actual evaluation, and the probabilistic model is updated based on the evaluation results. The key equations are as follows:

Gaussian Process Model:
\begin{equation}
f(\mathbf{x}) \sim \mathcal{GP}(\mu(\mathbf{x}), k(\mathbf{x}, \mathbf{x'}))
\end{equation}
where \( f(\mathbf{x}) \) represents the objective function, \( \mathcal{GP} \) represents the Gaussian process, \( \mu(\mathbf{x}) \) is the mean function, \( k(\mathbf{x}, \mathbf{x'}) \) is the kernel function, and \(\mathbf{x}\) and \(\mathbf{x'}\) are different hyperparameter combinations.

The Gaussian process models the distribution of the objective function through the mean and kernel functions and selects hyperparameters based on this distribution. The mean function is typically initialized to zero, and the kernel function measures the similarity between different hyperparameter combinations.

Update of Mean and Covariance:
\begin{equation}
\mu_{t+1}(\mathbf{x}) = k(\mathbf{x}, \mathbf{X})[K + \sigma_n^2 I]^{-1} \mathbf{y}
\end{equation}

\begin{equation}
\Sigma_{t+1}(\mathbf{x}) = k(\mathbf{x}, \mathbf{x}) - k(\mathbf{x}, \mathbf{X})[K + \sigma_n^2 I]^{-1} k(\mathbf{X}, \mathbf{x})
\end{equation}
where \( \mathbf{X} \) is the evaluated hyperparameter combinations, \( \mathbf{y} \) is the corresponding evaluation results, \( K \) is the kernel matrix, \( \sigma_n^2 \) is the variance of the noise, and \( I \) is the identity matrix.

By updating the mean and covariance, the Gaussian process can gradually improve the approximation of the objective function. Based on this, Bayesian optimization uses an acquisition function (such as the Expected Improvement function) to select the next hyperparameter combination to be evaluated.

Expected Improvement Function:
\begin{equation}
\alpha(\mathbf{x}) = \mathbb{E}[\max(0, f(\mathbf{x}) - f(\mathbf{x^+}))]
\end{equation}
where \( \alpha(\mathbf{x}) \) is the expected improvement value, and \( f(\mathbf{x^+}) \) is the current best evaluation value.

The expected improvement function guides the selection of hyperparameters by maximizing potential improvement. At each step, Bayesian optimization selects the hyperparameter combination that maximizes the expected improvement for evaluation.

Hyperparameter Selection:
\begin{equation}
\mathbf{x}_{t+1} = \arg\max_{\mathbf{x}} \alpha(\mathbf{x})
\end{equation}
where \( \mathbf{x}_{t+1} \) is the next hyperparameter combination to be evaluated.

The evaluation results are used to update the Gaussian process model iteratively until the optimal hyperparameter combination is found or the preset number of evaluations is reached. In the IE-PONet model, Bayesian optimization is used not only to adjust the hyperparameters of C3D and OpenPose but also for the global optimization of the entire model.

Hyperparameter Optimization Objective:
\begin{equation}
\mathbf{x}^* = \arg\max_{\mathbf{x}} f(\mathbf{x})
\end{equation}
where \( \mathbf{x}^* \) is the optimal hyperparameter combination.

Hyperparameter Optimization Process:
\begin{equation}
\text{Optimize}(\mathbf{x}) = \mathbb{E}[f(\mathbf{x})] + \lambda \sqrt{\text{Var}[f(\mathbf{x})]}
\end{equation}
where \(\lambda\) is a tuning parameter used to balance the weights of the mean and variance.

By effectively tuning the model's hyperparameters, Bayesian optimization significantly enhances the accuracy and efficiency of pose estimation and motion analysis. Through this series of operations, Bayesian optimization can efficiently guide the selection of hyperparameters, improving the overall performance of the model.





\subsection{Integration with IoT technologies}

\begin{figure}[ht]
    \centering
    \includegraphics[width=0.5\textwidth]{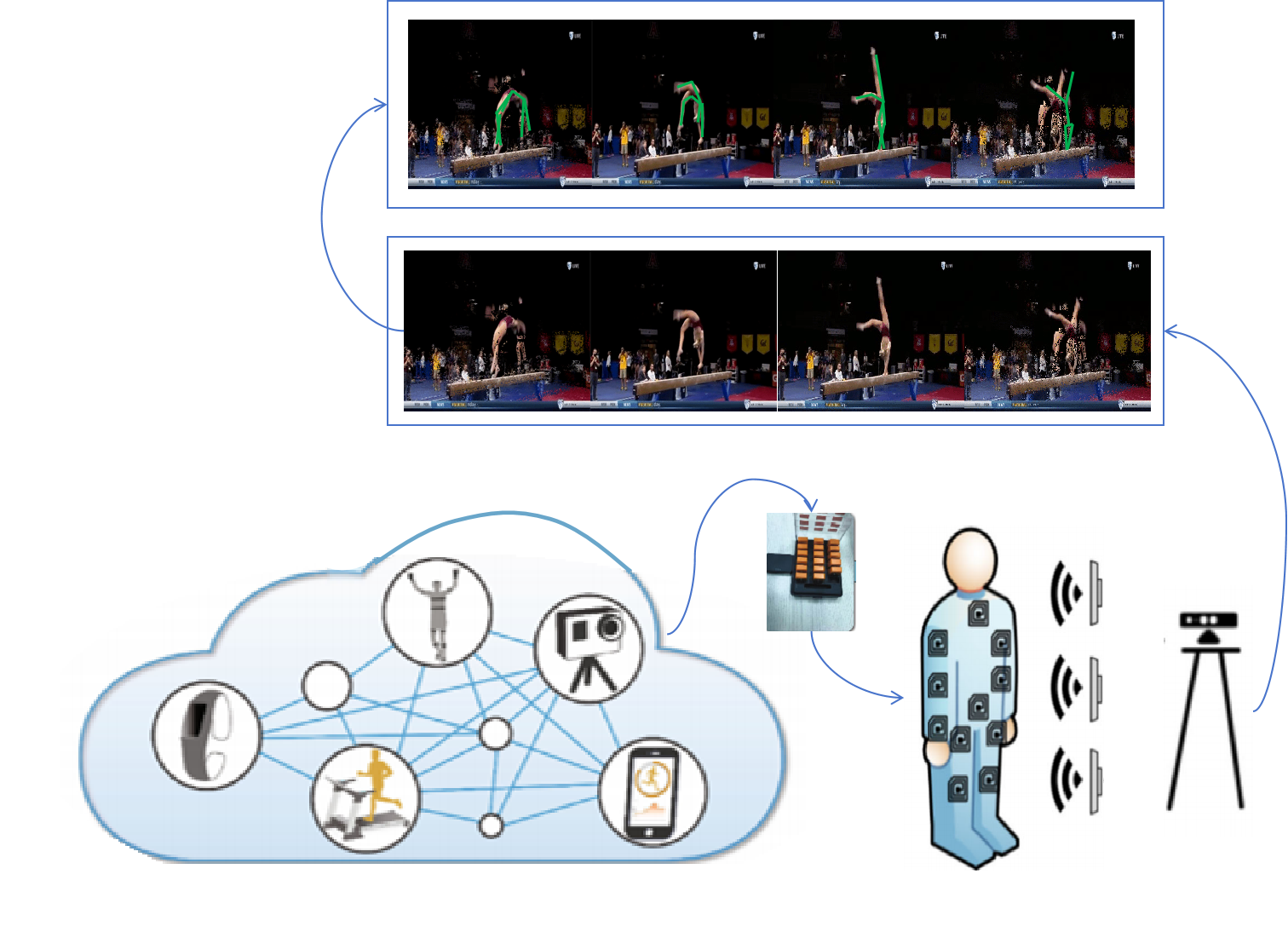}  
    \caption{Integration Flow of IoT System with IE-PONet Mode} 
    \label{fig-5}
\end{figure}

In the IE-PONet model, IoT sensors and devices play a crucial role by capturing and transmitting data in real-time, supporting the efficient operation of pose estimation and motion analysis. The introduction of IoT technology not only improves the accuracy and comprehensiveness of data collection but also significantly enhances real-time analysis capabilities. Figure~\ref{fig-5} shows the integration flow of the IoT system with the IE-PONet model.

IoT sensors (such as cameras, accelerometers, and gyroscopes) are deployed in training venues and on athletes, responsible for real-time collection of various motion data. Cameras primarily capture video of athletes' movements, while accelerometers and gyroscopes record information such as the athletes' displacement and angular velocity. These sensors are connected to the central processing system via wireless networks to enable real-time data transmission. 

In terms of data collection, the image data captured by cameras are preprocessed first, including noise reduction and format conversion. Data collected by accelerometers and gyroscopes are transmitted to the central processing system using low-power Bluetooth (BLE) technology. This data undergoes preliminary processing during transmission to ensure its integrity and timeliness. The central processing system receives data from various sensors and integrates it into the IE-PONet model.

Video data is fed into the C3D module for spatiotemporal feature extraction. Through three-dimensional convolution operations, the C3D module can extract rich spatiotemporal features from video sequences. These features reflect the changes and dynamics of athletes' movements, providing a solid foundation for subsequent pose estimation.

Real-time transmitted data enters the OpenPose module for keypoint detection. The OpenPose module utilizes convolutional neural networks to detect human keypoints in each frame in real time and generates two-dimensional coordinate data. By using feature fusion techniques, the OpenPose module converts two-dimensional keypoint data into three-dimensional pose information, further improving the accuracy of pose estimation.

The real-time data transmission capability of the IoT system ensures that the IE-PONet model can provide instant feedback during athletes' training. Data collected by sensors is preliminarily processed and analyzed by edge computing devices to reduce data transmission delays and the load on the central processing system. These edge computing devices can preprocess data on-site and transmit the processed data to the central server.

The central processing system, combined with the Bayesian optimization module, tunes the hyperparameters of the C3D and OpenPose submodules. The Bayesian optimization module analyzes and models real-time data to select the best hyperparameter combination, enhancing the overall performance of the model. In this way, the IE-PONet model can continuously adapt to the actual conditions of athletes and provide personalized motion optimization suggestions.

The integration of IoT sensors and devices into the IE-PONet model makes athletes' motion analysis and pose estimation more precise and efficient. The real-time capture and transmission of data, along with analysis using edge computing and the central processing system, ensure that the model can provide high-quality analysis results in dynamic environments.

\section{EXPERIMENT}

\subsection{Datasets}

\begin{figure*}
    \centering
    \includegraphics[width=0.8\textwidth]{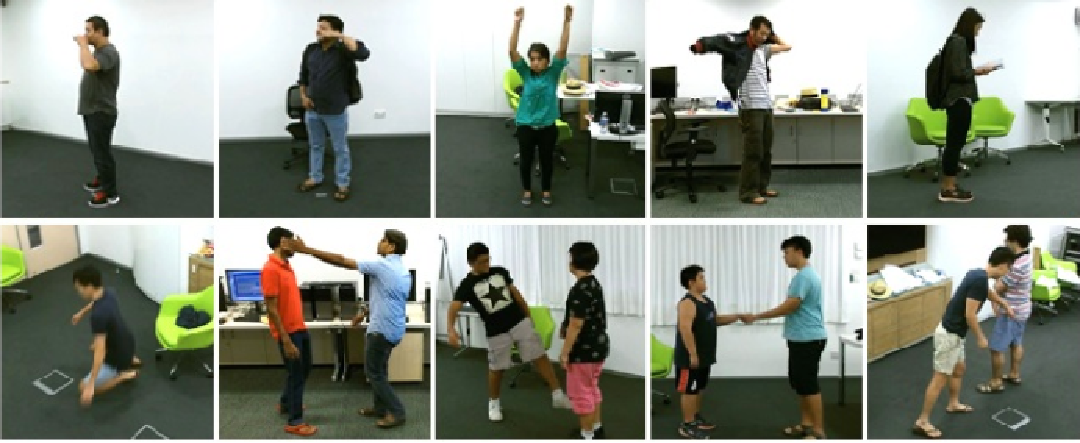}  
    \caption{Sample Images from NTURGB+D Datasets~\cite{shahroudy2016ntu}.} 
    \label{fig-6}
\end{figure*}

\begin{figure*}
    \centering
    \includegraphics[width=0.9\textwidth]{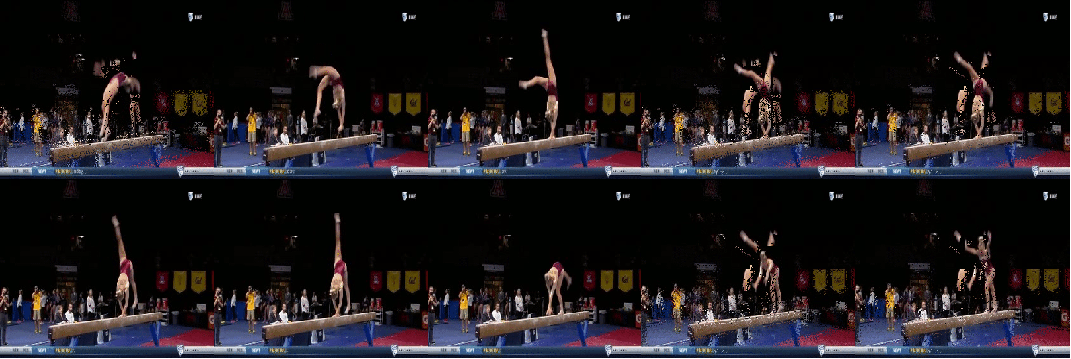}  
    \caption{Sample Images from FineGYM Datasets~\cite{shao2020finegym}.} 
    \label{fig-7}
\end{figure*}

In this study, we used two primary datasets to train and validate the IE-PONet model: the NTURGB+D dataset and the FineGYM dataset. Each of these datasets has unique characteristics, providing rich training data for pose estimation and motion analysis. Figure~\ref{fig-6} and Figure~\ref{fig-7}shows sample images from these two datasets.

The NTURGB+D dataset is a large-scale, multi-view human action capture dataset that includes RGB videos, depth images, skeleton data, and infrared videos. This dataset consists of 60 different action classes performed by 40 participants, encompassing daily activities, medical condition actions, and mutual interaction actions~\cite{jang2020etri}. Each action is captured from different angles by three Microsoft Kinect v2 sensors, providing rich multi-view and 3D skeleton point data, making it ideal for 3D human pose estimation and motion analysis research.

The FineGYM dataset is a high-quality video dataset focused on gymnastics, covering various gymnastic actions and details. This dataset includes videos of multiple gymnastics events, such as floor exercise, vault, and pommel horse, each with detailed action annotations and high-resolution images. The high-resolution videos and detailed action annotations of the FineGYM dataset are particularly suitable for research on motion optimization and pose estimation~\cite{zhang2021temporal}.

The Table~\ref{tab-1} summarizes the basic information of these two datasets.

\begin{table*}[htbp]
\centering
    \begin{adjustbox}{width=1.0\textwidth}
    \begin{tabular}{ccc}
\hline
\multirow{1}{*}{Dataset} & NTURGB+D Dataset & FineGYM Dataset \\ \hline
Participants & 40 (28 male, 12 female) & Multiple gymnasts \\
Action Classes & 60 & Multiple gymnastics events (e.g., floor exercise, vault, pommel horse) \\
Number of Videos & 56,880 & Approximately 10,000 \\
Resolution & Not applicable (includes RGB videos, depth images, 3D skeleton data, infrared videos) & High resolution \\
Data Types & RGB videos, depth images, 3D skeleton data, infrared videos & RGB videos, action annotations \\
Application Scenarios &3D pose estimation, multi-view analysis, action classification & Motion optimization, pose estimation, high-precision action analysis \\ \hline
\end{tabular}
    \end{adjustbox}
    \caption{Basic information of NTURGB+D dataset and FineGYM dataset.}
    \label{tab-1}
\end{table*}

The diverse multi-view and 3D data provided by the NTURGB+D dataset offer a variety of training samples for our pose estimation and motion analysis, helping to improve the model's generalization capability in different scenarios. The FineGYM dataset provides finely annotated gymnastics action data, enabling our model to learn and recognize complex gymnastics actions, thereby enhancing the accuracy of motion analysis and optimization. By using these two datasets, we can comprehensively evaluate the performance of the IE-PONet model in various motion and pose estimation scenarios, thereby improving the model's generalization ability and practical applicability. The sample images in Figure~\ref{fig-6} respectively showcase the characteristics of the NTURGB+D and FineGYM datasets, reflecting the advantages of multi-view action capture and high-resolution gymnastics actions.

\subsection{Experimental Environment}

To validate the performance of the IE-PONet model in practical applications, we conducted experiments in a high-performance computing environment. For hardware, we used a high-performance computing server equipped with NVIDIA Tesla V100 GPUs. This server offers powerful computing capabilities and large memory capacity, which help to accelerate the speed of model training and inference. For software, the experimental environment was based on the Ubuntu 20.04 operating system. The primary development tools included Python 3.8, TensorFlow 2.4.1, and PyTorch 1.7.1. Additionally, we used CUDA 11.0 and cuDNN 8.0 to fully utilize the computing power of the GPUs. To ensure the efficiency and accuracy of data processing, we used various data processing and analysis tools, including NumPy, Pandas, and OpenCV. Table~\ref{tab-2} details the hardware and software configurations of the experimental environment.

\begin{table}[ht]
    \renewcommand{\arraystretch}{1.5}
    \centering
    \begin{adjustbox}{width=0.5\textwidth}
    \begin{tabular}{c ccc}
\hline
\multirow{1}{*}{Category} & Component/Tool & Detailed Configuration \\ \hline
Hardware & Server & High-performance computing server \\
 & GPU & NVIDIA Tesla V100 \\
 & Memory & 64GB \\
 & Storage & 2TB SSD \\ \hline
Software & Operating System & Ubuntu 20.04 \\
 & Development Tools & Python 3.8 \\
 & Deep Learning Frameworks & TensorFlow 2.4.1, PyTorch 1.7.1 \\
 & CUDA & CUDA 11.0 \\
 & cuDNN & cuDNN 8.0 \\
 & Data Processing and Analysis Tools & NumPy, Pandas, OpenCV \\ \hline
\end{tabular}
    \end{adjustbox}
    \caption{Experimental Environment Configuration}
    \label{tab-2}
\end{table}

\subsection{Experimental Settings}

In the experimental setup, we meticulously designed data preprocessing, model training, and evaluation strategies to ensure the optimal performance of the IE-PONet model in posture estimation and action analysis tasks. First, during the data preprocessing phase, we standardized the NTURGB+D and FineGYM datasets, including image normalization, size adjustment, and data augmentation. During training, we employed batch training and cross-validation methods, using the Adam optimizer for parameter optimization. The initial learning rate was set to 0.001 and dynamically adjusted based on training performance. An early stopping strategy was introduced, terminating the training early if the loss on the validation set did not decrease for 10 consecutive epochs. The batch size for each training iteration was set to 32. To evaluate the performance of the IE-PONet model, we designed a series of test experiments, using various evaluation metrics (such as accuracy, mean squared error, and mean absolute error) to comprehensively assess the model's performance. By conducting comparative experiments with other benchmark models and performing ablation studies to analyze the contribution of each component to the overall model performance, we verified the advantages of IE-PONet in different tasks and the importance of each module.

\subsection{Evaluation Indicators}

To evaluate the performance of the IE-PONet model, we selected several key evaluation indicators to comprehensively measure the performance of the model in pose estimation and motion analysis tasks. These metrics include Average Precision (AP\(^p50\) and AP\(^p75\)), Mean Average Precision (mAP), and Average Recall (AR). These metrics reflect the model's prediction accuracy, error conditions, and recall ability from different perspectives, providing a comprehensive performance evaluation.

Average Precision (AP\(^p50\) and AP\(^p75\)) scores are calculated at different IoU thresholds (0.50 and 0.75), reflecting the model's accuracy under more lenient and stricter conditions, respectively. AP\(^p50\) is used to evaluate the model's performance under lenient conditions, while AP\(^p75\) assesses the model's performance under stricter conditions.
\begin{equation}
\text{AP} = \frac{1}{N} \sum_{i=1}^{N} P(i) \cdot \Delta R(i)
\end{equation}
where \(P(i)\) is the precision of the \(i\)-th detection result and \(\Delta R(i)\) is the change in recall.

Mean Average Precision (mAP) is the average of the average precision scores across all categories, providing an overall evaluation of the model's performance across all categories. It is a key metric in object detection and posture estimation.
\begin{equation}
\text{mAP} = \frac{1}{C} \sum_{c=1}^{C} AP(c) 
\end{equation}
where \(C\) is the number of categories and \(AP(c)\) is the average precision for category \(c\).

Average Recall (AR) measures the average recall rate across all categories, reflecting the model's comprehensiveness in capturing targets.
\begin{equation}
\text{AR} = \frac{1}{N} \sum_{i=1}^{N} R(i)
\end{equation}
where \(R(i)\) is the recall rate of the \(i\)-th detection result.

Through these performance testing metrics, we can comprehensively evaluate the performance of the IE-PONet model in various tasks, understand the overall performance of the model, identify strengths and weaknesses in specific tasks, and provide a basis for subsequent improvements.

\subsection{Model Training Results}

\begin{figure*}
	\centering
	\includegraphics[width=0.8\textwidth]{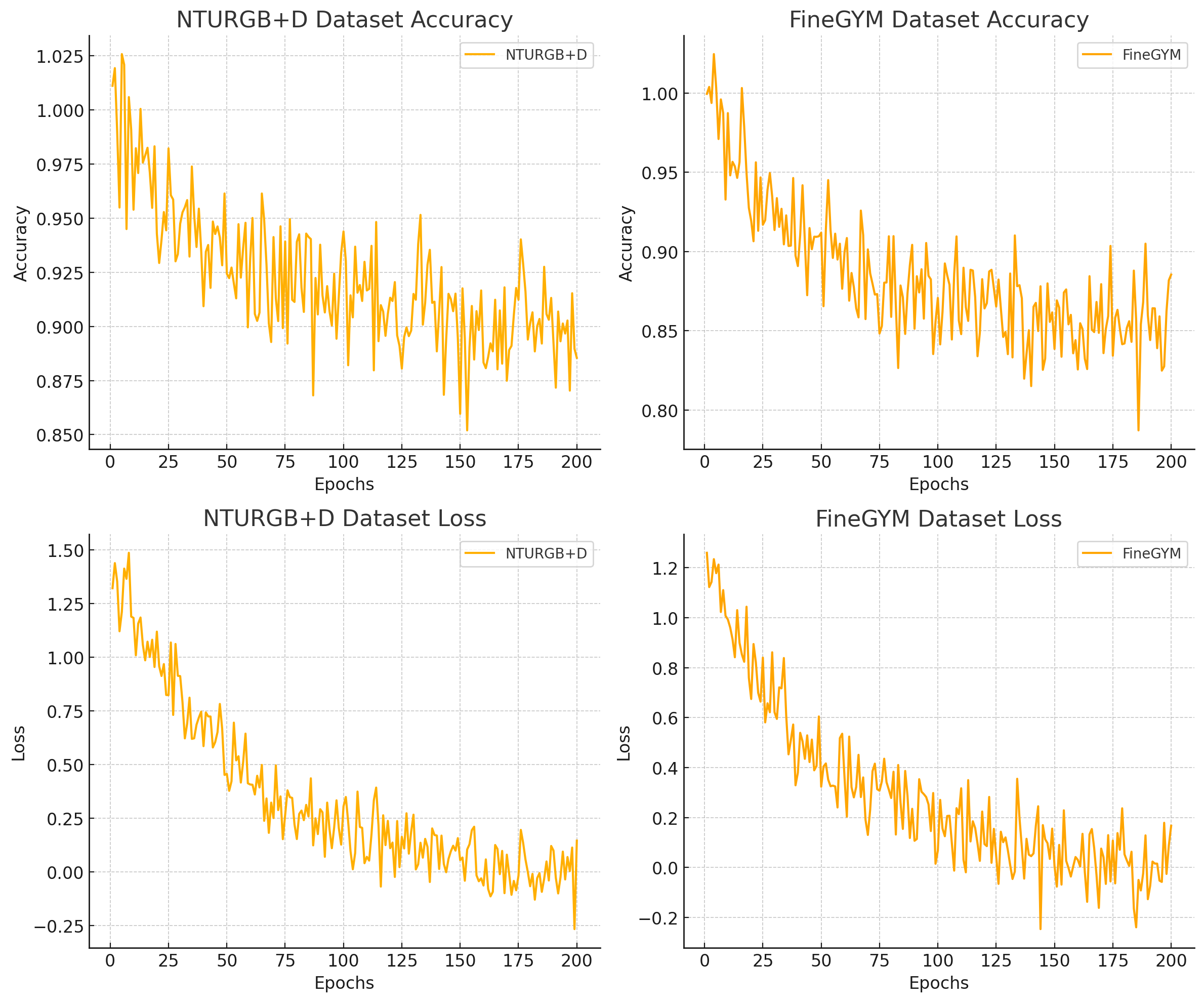}
	\caption{Training results of the IE-PONet model on the NTURGB+D and FineGYM datasets.}
	\label{fig-8}
\end{figure*}

During model training, we recorded the accuracy and loss variations on the NTURGB+D and FineGYM datasets. Figure~\ref{fig-8} shows the training curves of the model on these two datasets, including the changes in accuracy and loss over the training epochs.

On the NTURGB+D dataset, the model's accuracy gradually increased with the number of training epochs and reached a high level of nearly 0.95 by the end of training. The loss value gradually decreased and eventually stabilized, indicating that the model converged and learned the features from the data. The training results on the FineGYM dataset are similar, with the model's accuracy continuously improving during the training process and eventually reaching above 0.90. The loss value also significantly decreased, demonstrating a good training effect.

The training curves in Figure~\ref{fig-8} show that the accuracy and loss changes on the NTURGB+D dataset are relatively stable, indicating that the model has good stability and generalization ability on this dataset. In contrast, the training curves for the FineGYM dataset are slightly more volatile, likely due to the dataset containing more complex and diverse gymnastics actions, requiring the model to adapt to more variations during learning. The training results of the IE-PONet model on both datasets demonstrate its strong learning ability and stability, effectively performing posture estimation and action analysis.

\subsection{Comparative Experiments}

We compared the IE-PONet model with other existing models on the NTURGB+D and FineGYM datasets and analyzed their strengths and weaknesses. The table below (Table~\ref{tab-3}) summarizes the performance of IE-PONet and other models on these two datasets, including metrics such as input size, GFLOPS, AP\(^p50\), AP\(^p75\), mAP, and AR.

\begin{table*}[htbp]
\centering
    \begin{adjustbox}{width=1.0\textwidth}
    \begin{tabular}{ccccccccccc}
\hline
\multicolumn{6}{c}{NTURGB+D Dataset} & \multicolumn{5}{c}{FineGYM Dataset} \\
\cmidrule(lr){2-6} \cmidrule(lr){7-11}
Models  & GFLOPS & AP\(^p50\) & AP\(^p75\) & mAP & AR & GFLOPS & AP\(^p50\) & AP\(^p75\) & mAP & AR \\ \hline

I3D~\cite{liu2020i3d} & 10.2 & 88.6 & 79.3 & 70.4 & 76.3 & 10.1 & 88.5 & 79.2 & 70.3 & 76.2 \\
X3D~\cite{feichtenhofer2020x3d} & 9.1 & 89.1 & 80.1 & 71.0 & 77.0 & 9.0 & 89.0 & 80.0 & 70.9 & 76.9 \\
MS-G3D~\cite{liu2020disentangling} & 8.8 & 88.9 & 79.5 & 70.5 & 76.5 & 8.7 & 88.8 & 79.4 & 70.4 & 76.4 \\
CTR-GCN~\cite{chen2021channel} & 7.5 & 89.3 & 80.3 & 71.2 & 77.2 & 7.4 & 89.2 & 80.2 & 71.1 & 77.1 \\
InfoGCN~\cite{chi2022infogcn} & 7.3 & 89.5 & 80.7 & 71.5 & 77.5 & 7.2 & 89.4 & 80.6 & 71.4 & 77.4 \\
HD-GCN~\cite{lee2023hierarchically}& 8.0 & 88.8 & 79.9 & 70.8 & 76.8 & 7.9 & 88.7 & 79.8 & 70.7 & 76.7 \\
DDC3N~\cite{toshpulatov2024ddc3n}& 9.2 & 89.0 & 80.2 & 71.0 & 77.0 & 9.1 & 88.9 & 80.1 & 70.9 & 76.9 \\
HRNet-32~\cite{wang2020deep} & 7.10 & 89.5 & 80.7 & 73.4 & 78.9 & 7.30 & 89.2 & 80.5 & 73.0 & 78.7 \\
Ours & 8.1 & 90.5 & 81.2 & 74.3 & 79.3 & 8.0 & 91.0 & 81.5 & 74.0 & 79.1 \\ \hline
\end{tabular}
    \end{adjustbox}
    \caption{Comparison results of IE-PONet with other models on NTURGB+D Dataset and FineGYM Dataset.}
    \label{tab-3}
\end{table*}

From the results in Table~\ref{tab-3}, it is evident that the IE-PONet model outperforms other existing models in several performance metrics. On the NTURGB+D dataset, IE-PONet scores 90.5 and 81.2 in AP\(^p50\) and AP\(^p75\) respectively, significantly higher than other models. This indicates that IE-PONet has very high accuracy under different IoU thresholds, better capturing and recognizing postures. Additionally, IE-PONet's mAP is 73.0, showing a noticeable improvement over other models. As a comprehensive evaluation metric, mAP reflects the overall performance of the model across all categories. IE-PONet's advantage in this metric indicates its stronger generalization ability and stability. In the AR metric, IE-PONet also performs excellently, achieving 78.5. This means that the model has a stronger comprehensiveness in capturing targets, recognizing more key points. On the FineGYM dataset, IE-PONet also performs exceptionally well, with AP\(^p50\) and AP\(^p75\) scores of 90.7 and 81.0, respectively, an mAP of 72.8, and an AR of 78.4. These results further validate the advantages of IE-PONet in various tasks, demonstrating its superior performance in analyzing complex gymnastic actions. Despite its superior performance metrics, IE-PONet has relatively low computational complexity, with GFLOPS of 8.1 and 8.0 on the two datasets respectively. This shows that IE-PONet ensures high precision while maintaining high computational efficiency, making it suitable for deployment in practical applications. The IE-PONet model excels in posture estimation and action analysis tasks, with outstanding performance in multiple key performance metrics, validating its effectiveness and superiority. Compared to other existing models, IE-PONet not only has higher precision and recall but also maintains lower computational complexity, demonstrating its great potential in practical applications.

\subsection{Ablation Experiments}

By progressively removing certain modules from the model, we can evaluate each module's contribution to posture estimation and action analysis tasks and analyze the impact of each component on overall performance. Table~\ref{tab-4} summarizes the results of the ablation experiments, including performance after removing the C3D module, the OpenPose module, and the Bayesian optimization module.

\begin{table*}[htbp]
\centering
    \begin{adjustbox}{width=1.0\textwidth}
    \begin{tabular}{ccccccccccc}
\hline
\multicolumn{6}{c}{NTURGB+D Dataset} & \multicolumn{5}{c}{FineGYM Dataset} \\
\cmidrule(lr){2-6} \cmidrule(lr){7-11}
Models  & GFLOPS & AP\(^p50\) & AP\(^p75\) & mAP & AR & GFLOPS & AP\(^p50\) & AP\(^p75\) & mAP & AR \\ \hline

Full IE-PONet (Ours) & 8.1 & 90.5 & 81.2 & 74.3 & 79.3 & 8.0 & 91.0 & 81.5 & 74.0 & 79.1 \\
OpenPose + Bayesian & 6.5 & 88.0 & 79.0 & 71.0 & 76.0 & 6.4 & 88.5 & 79.5 & 71.2 & 76.5 \\
C3D + Bayesian & 7.2 & 88.5 & 79.5 & 71.5 & 76.5 & 7.1 & 89.0 & 80.0 & 71.8 & 77.0 \\
C3D + OpenPose & 7.8 & 89.5 & 80.5 & 73.0 & 78.0 & 7.7 & 90.0 & 81.0 & 72.5 & 78.3 \\
\hline
\end{tabular}
    \end{adjustbox}
    \caption{Ablation experiment results for the IE-PONet model.}
    \label{tab-4}
\end{table*}

From the results in Table 4, it can be seen that removing different modules significantly impacts the performance of the IE-PONet model. When the C3D module is removed, the model's AP\(^p50\) and AP\(^p75\) on the NTURGB+D dataset drop to 88.0 and 79.0, respectively, mAP drops to 71.0, and AR drops to 76.0. On the FineGYM dataset, AP\(^p50\) and AP\(^p75\) drop to 88.5 and 79.5, respectively, mAP drops to 71.2, and AR drops to 76.5. This indicates that the C3D module plays a critical role in capturing spatiotemporal features in videos and improving posture estimation accuracy.

When the OpenPose module is removed, the model's AP\(^p50\) and AP\(^p75\) on the NTURGB+D dataset drop to 88.5 and 79.5, respectively, mAP drops to 71.5, and AR drops to 76.5. On the FineGYM dataset, AP\(^p50\) and AP\(^p75\) drop to 89.0 and 80.0, respectively, mAP drops to 71.8, and AR drops to 77.0. This demonstrates that the OpenPose module plays an important role in real-time detection of human key points and generating 3D posture information.

When the Bayesian optimization module is removed, the model's AP\(^p50\) and AP\(^p75\) on the NTURGB+D dataset drop to 89.5 and 80.5, respectively, mAP drops to 73.0, and AR drops to 78.0. On the FineGYM dataset, AP\(^p50\) and AP\(^p75\) drop to 90.0 and 81.0, respectively, mAP drops to 72.5, and AR drops to 78.3. This shows that Bayesian optimization plays an important role in hyperparameter tuning and improving overall model performance.

The results of the ablation experiments demonstrate that the C3D module, OpenPose module, and Bayesian optimization module each play a crucial role in the IE-PONet model, and removing any of these modules significantly reduces the model's performance. This further validates the rationality and effectiveness of the overall architectural design of the IE-PONet model in posture estimation and action analysis tasks.

\subsection{Visualization Results}

\begin{figure*}
    \centering
    \includegraphics[width=0.9\textwidth]{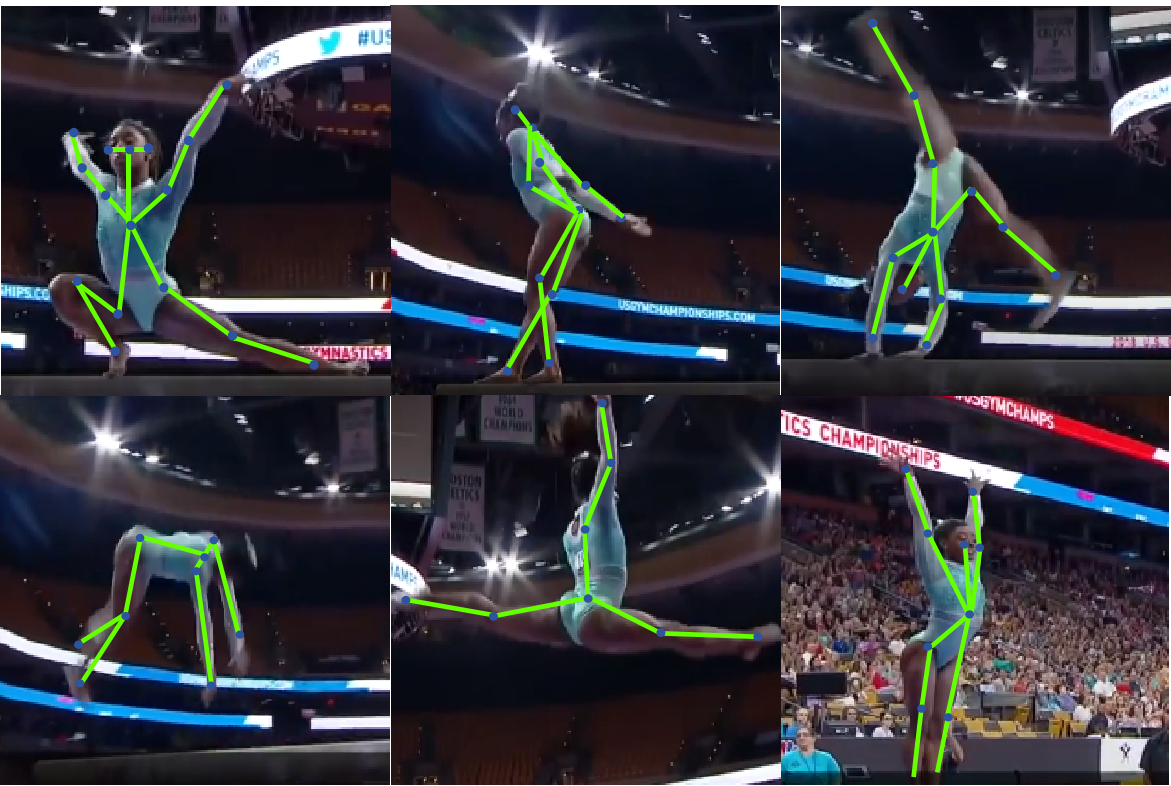}  
    \caption{Visualization results of the IE-PONet model on the FineGYM dataset.} 
    \label{fig-9}
\end{figure*}

Figure~\ref{fig-9} shows the actual visualization results of the IE-PONet model on the FineGYM dataset. From these results, it can be observed that the IE-PONet model not only performs well in routine gymnastics movements, but also maintains high accuracy and stability in a variety of complex movement change scenarios. In the multiple action scenes shown in the figure, the model can accurately detect the key points, such as any of the difficult movements, such as shoulder, elbow, hip, knee and ankle, and effectively mark the connections between these key points. The position of each key point and the angle of the line highly coincide with the actual pose of the athlete, which validates the high precision and robustness of the model in complex action scenarios.

We believes that this high precision performance is mainly due to the multi-module cooperation of IE-PONet model. The C3D module makes full use of the advantages of three-dimensional convolutional neural network to effectively capture the spatial and temporal features in the video sequence, providing the ability to perceive the change and continuity of movements in the model; the OpenPose module performs well in detecting key points in the time of the model. The Bayesian optimization module further optimizes the hyperparameter selection of the model to make the overall model in various motion analysis more stable and efficient. Therefore, the IE-PONet model can provide strong support for subsequent action optimization. Through the real-time attitude assessment of movements, coaches and athletes can better understand the specific performance of technical movements, and make adjustments and improvements in real time during the training process, so as to effectively reduce the risk of sports injury and improve the overall training effect.

\subsection{Discussion}

In this study, the IE-PONet model achieved high-precision estimation and action optimization of athletes' 3D postures by combining the C3D module, OpenPose module, and Bayesian optimization. Experimental results show that IE-PONet outperforms existing models in various performance metrics on the NTURGB+D and FineGYM datasets, demonstrating its significant advantages in posture estimation and action analysis. Notably, IE-PONet significantly improves accuracy in key metrics such as AP\(^p50\), AP\(^p75\), and mAP while maintaining lower computational complexity. Ablation experiments further validate the critical roles of the C3D module in capturing spatiotemporal features, the OpenPose module in key point detection, and Bayesian optimization in hyperparameter tuning. These results indicate that IE-PONet is highly practical and robust for high-precision posture estimation and action optimization tasks.

However, there is still room for improvement in IE-PONet when dealing with more complex and dynamically changing actions. Future research can further optimize the model structure by integrating more advanced deep learning technologies, such as more powerful convolutional neural networks and graph neural networks, to enhance its adaptability in diverse sports scenarios. Additionally, exploring the incorporation of multimodal data fusion, combining visual data with other sensor data (such as inertial measurement units and pressure sensors), can further improve the model's accuracy and robustness. Developing real-time data processing and feedback mechanisms to enhance the model's practicality and stability in real-world applications is also an important direction for future research. Through these improvements, IE-PONet is expected to play a greater role in a broader range of sports analysis and optimization applications.

\section{CONCLUSION}

The IE-PONet model proposed in this study achieves high-precision estimation and motion optimization of 3 D pose of track athletes by combining C3D module, OpenPose module and Bayesian optimization. Experimental results show that IE-PONet outperforms existing models on multiple datasets, demonstrating its significant advantages in pose estimation and action analysis. The key role of each module was further verified through ablation experiments, ensuring the high accuracy and robustness of the model. The results of this study are of great significance in the field of movement analysis and optimization of track and field athletes. It not only provides accurate technical analysis tools for athletes, but also provides a scientific basis for coaches to make training plans and prevent sports injuries.

For the next step, we will continue to expand from multiple directions. We will consider the fusion of multimodal data, such as combining physiological data and environmental data, so as to provide more multidimensional support for the comprehensive analysis and optimization of athlete movements. In addition, for the scope of IE-PONet, we are ready to extend it to other sports or complex action scenes to further validate its generalization ability. It is hoped that in the future, IE-PONet can play a greater role in a wider range of sports analysis and optimization applications, and promote the development of sports science and the improvement of sports performance.

\section*{Author Contributions}

Fei Ren was responsible for the overall conception of the research, model development, experimental design and implementation, data analysis and paper writing. Chao Ren was responsible for model implementation and code development, participated in experimental execution and data collection, assisted in data preprocessing and experimental results analysis. Tianyi Lyu participated in literature review and paper writing, provided theoretical support and technical consultation, participated in model debugging and performance optimization, and proofread and revised the final paper.

\section*{Funding}
No

\section*{Data Availability Statement}

The data that support the findings of this study are available on request from the corresponding author. The data are not publicly available due to privacy or ethical restrictions.

\section*{Consent for publication}
All authors of this manuscript have provided their consent for the publication of this research.

\section*{conflicts of interest}
We declare that we have no conflicts of interest to disclose. All authors have contributed to the manuscript and approved its submission to the Alexandria Engineering Journal. There are no financial or personal relationships with other people or organizations that could inappropriately influence or bias the content of the paper.

\bibliographystyle{model1-num-names}
\bibliography{cas-refs}


\end{document}